\def\year{2022}\relax
\documentclass[letterpaper]{article} % DO NOT CHANGE THIS
\usepackage{aaai22}  % DO NOT CHANGE THIS
\usepackage{times}  % DO NOT CHANGE THIS
\usepackage{helvet}  % DO NOT CHANGE THIS
\usepackage{courier}  % DO NOT CHANGE THIS
\usepackage[hyphens]{url}  % DO NOT CHANGE THIS
\usepackage{graphicx} % DO NOT CHANGE THIS
\usepackage{natbib}  % DO NOT CHANGE THIS AND DO NOT ADD ANY OPTIONS TO IT
\usepackage{caption} % DO NOT CHANGE THIS AND DO NOT ADD ANY OPTIONS TO IT
\DeclareCaptionStyle{ruled}{labelfont=normalfont,labelsep=colon,strut=off} % DO NOT CHANGE THIS
\usepackage{algorithm}
\usepackage{algorithmic}
\usepackage{booktabs}
\usepackage{multirow}
\usepackage{tabularx}
\usepackage{newfloat}
\usepackage{listings}
\floatstyle{ruled}
\newfloat{listing}{tb}{lst}{}
\floatname{listing}{Listing}
\title{Hierarchical Cross-Modality Semantic Correlation Learning Model \\for Multimodal Summarization}
\author {
    Litian Zhang,\textsuperscript{1,\dag}
    Xiaoming Zhang,\textsuperscript{1,}\footnote{Corresponding author.}
    Junshu Pan,\textsuperscript{1,}\footnote{Equal contribution.}
    Feiran Huang\textsuperscript{2}
}
\affiliations {
    \textsuperscript{1}School of Cyber Science and Technology, Beihang University, Beijing, China\\
    \textsuperscript{2}College of Cyber Security, Jinan University, Guangzhou, China
\\
    \{litianzhang, 
    yolixs, 
    junshupan\}@buaa.edu.cn, huangfr@ieee.org
}

\usepackage{bibentry}
\newcommand{\tabincell}[2]{\begin{tabular}{@{}#1@{}}#2\end{tabular}} 
\begin{document}

\maketitle

\begin{abstract}
Multimodal summarization with multimodal output (MSMO) generates a summary with both textual and visual content. Multimodal news report contains heterogeneous contents, which makes MSMO nontrivial. Moreover, it is observed that different modalities of data in the news report correlate hierarchically. Traditional MSMO methods indistinguishably handle different modalities of data by learning a representation for the whole data, which is not directly adaptable to the heterogeneous contents and hierarchical correlation. In this paper, we propose a hierarchical cross-modality semantic correlation learning model (HCSCL) to learn the intra- and inter-modal correlation existing in the multimodal data. HCSCL adopts a graph network to encode the intra-modal correlation. Then, a hierarchical fusion framework is proposed to learn the hierarchical correlation between text and images. Furthermore, we construct a new dataset with relevant image annotation and image object label information to provide the supervision information for the learning procedure. Extensive experiments on the dataset show that HCSCL significantly outperforms the baseline methods in automatic summarization metrics and fine-grained diversity tests.

\end{abstract}

\section{Introduction}
\noindent With the rapid development of multimedia data on the Internet, multimodal summarization is a research direction worthy of attention and has broad development prospects. Therefore, some researches (\citealp{11zhu2018msmo,6li2020vmsmo,10zhu2020multimodal}) focus on studying multimodal summarization with multimodal output (MSMO) to help readers improve reading efficiency and satisfaction. However, these approaches abstract the summary from the raw data directly, which is ineffective in learning the latent and vital information from both the text and image content.

\begin{figure}
    % \centering
    \begin{center}
    \includegraphics[scale=0.24]{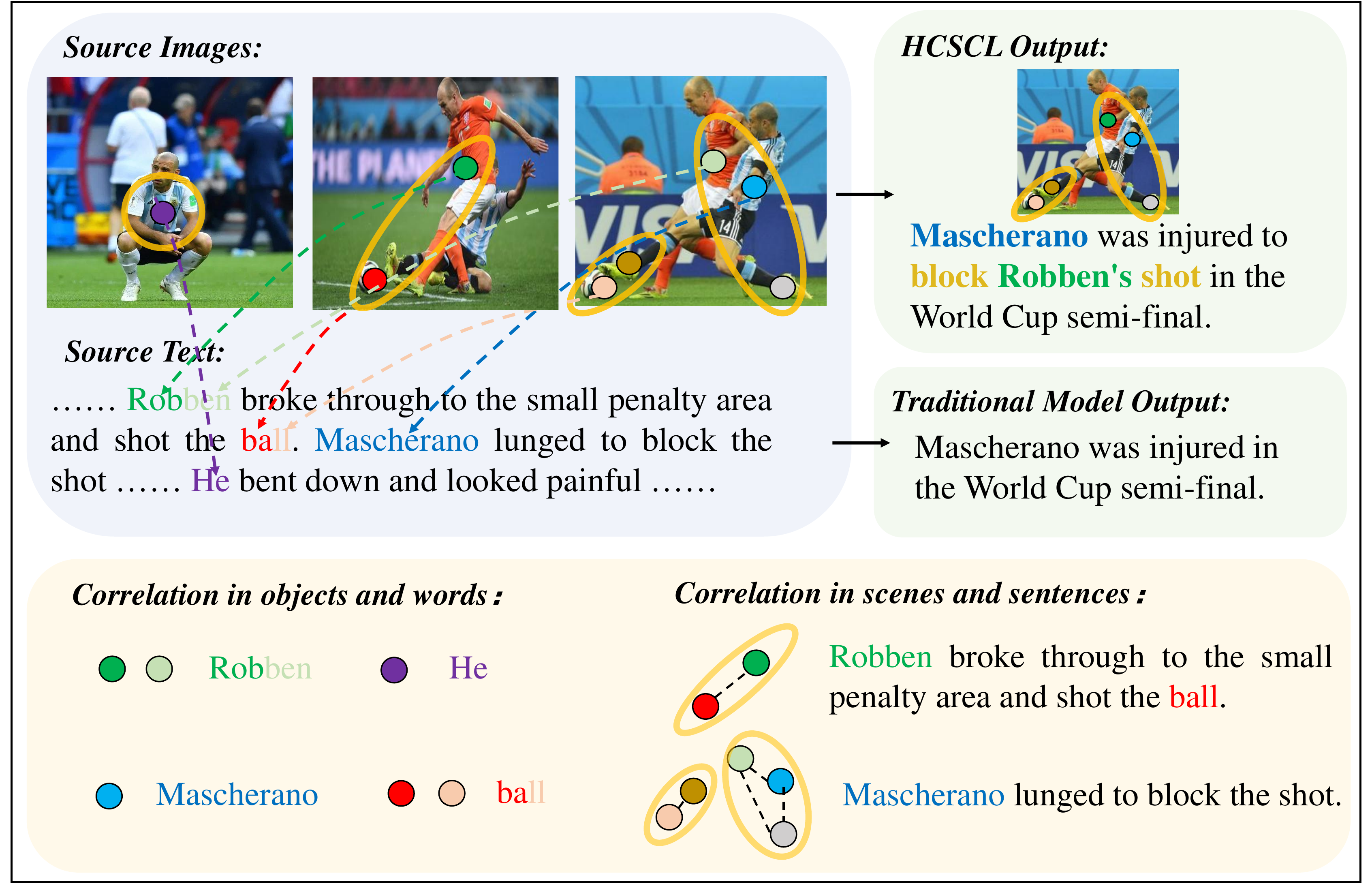}
    \end{center}
    \caption{An example of the cross-modality semantics correlation in multimodal data.}
    \label{fig:f1}
\end{figure}

Usually, the visual image and text article have heterogeneous structures. Directly mapping visual input and textual input as global vectors (\citealp{11zhu2018msmo,6li2020vmsmo,10zhu2020multimodal}) is not effective to learn the important information for both modalities from each other, and even noisy information is added to decrease the performance of summarization. Previous experiments (\citealp{11zhu2018msmo}) have shown that multimodal input models may decrease summarization metric scores compared to text-only input models. Our experiments also show that some multimodal input model methods perform worse than traditional text-only input models. Therefore, one of the core problems of MSMO is how to effectively learn from each other modalities of data to obtain high-quality summaries.

Meanwhile, the correlation between the visual content and text article presents unique characteristics, providing clues to learning the important information from the two modalities complementarily to improve MSMO. As shown in Figure 1, the low-level objects in an image constitute the high-level semantics called scenes through the interaction between them. Therefore, by analyzing the objects and hence the scenes, we can know what the image describes from different levels. In the other data space, words are also the basic textual information in an article, while the combination of words, called sentences, present more abstract semantics information. Besides the intra-modal correlation, the semantics objects in the image and article are correlated in different levels. For example, in Figure \ref{fig:f1}, each person in the image may be related to a name in the article, and the football sport in the image is also described by a sentence in the article. By learning the inter-modal correlation, it can be known what is the important information in both of the modalities. Moreover, even there are some incomplete descriptions in one modality, it can be learned from the other modality by exploiting the inter-modal correlation. As shown in Figure \ref{fig:f1}, we can generate the more complete information about the relation between the player “Mascherano” and the event “block Robben's shot”. Therefore, by exploiting the hierarchical cross-modality correlation, we can extract the important information from both image and article more effectively.

However, there is still a great challenge to learn the hierarchical cross-modality correlation. First, different modalities have different feature spaces and structures among the elements. It is nontrivial to learn an effective representation to reflect both the different content and structure information. Second, much noisy information might exist, while some important information might be missed in one of the modalities. However, there is no explicit knowledge about the correlation between different modalities of data. 

To tackle the challenge, we propose a novel Hierarchical Cross-Modality Semantic Correlation Learning model (HCSCL) to learn the intra- and inter-modality correlation for MSMO. In particular, two modality encoders are proposed to learn the intra-modal correlation for image and article, respectively. Then, a hierarchical fusion framework is proposed to learn the hierarchical correlation between image and article. A hierarchical attention method is proposed to combine the different levels of features learned by the hierarchical fusion framework to generate the summary. Furthermore, we construct a new dataset\footnote{https://github.com/LitianD/HCSCL-MSDataset/} with relevant image annotation to provide the supervision information for the learning procedure. Extensive experiments on the dataset show that HCSCL significantly outperforms the baseline methods in automatic summarization metrics and fine-grained diversity tests.
%  and image object label information
Our main contributions are as follows:

\begin{itemize}
\item   We propose a hierarchical learning model HCSCL to learn the intra- and inter-modality correlation in the multimodal data. To the best of our knowledge, this is the first work that guides multimodal summarization by exploiting the fine-grained semantics and their correlation information inside the multimodal data.
\item We propose a multimodal visual graph learning method to capture the structure and the content information and reinforce the inter-modality interaction. 
\item We construct a large-scale multimodal summarization dataset with relevant image annotations and object labels to evaluate the performance of MSMO.
\end{itemize}

\begin{figure*}[htbp]
    \begin{center}
    \includegraphics[scale=0.46]{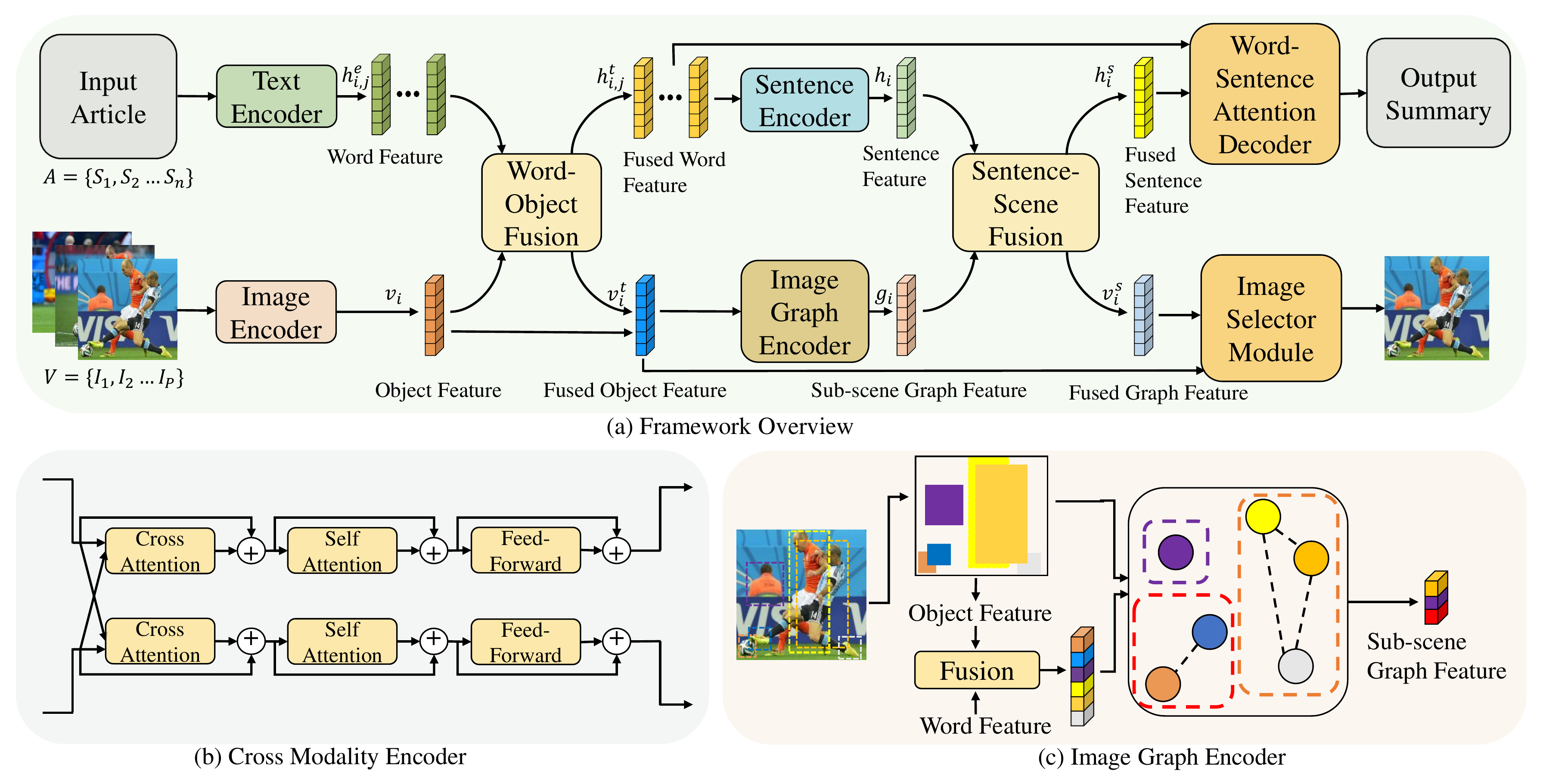}
    \end{center}
    \caption{(a) is the overview of HCSCL. It consists of three parts: Modality Feature Encoder, Hierarchical Semantic Relation Fusion, and Multimodal Output Summarizer. (b) is the Cross Modality Encoder used for hierarchical fusion. (c) is the Image Graph Encoder to connect relevant objects into scenes.}
    \label{fig:f2}
\end{figure*}

\section{Related works}
%应该体现数据集少 且缺少一些东西 扩展我们数据集好
\textbf{Multimodal Summarization with Text Only Output.} Unlike text summarization (\citealp{zhang2018neural}; \citealp{xiao2019extractive}; \citealp{see2017get}; \citealp{gao2019abstractive}), multimodal summarization (\citealp{1uzzaman2011multimodal}) is defined as a task to generate a condensed summary from a multimodal input, such as text, image, and video. Several works focus on generating a better text summary with the help of images (\citealp{3chen2018abstractive}; \citealp{12li2018multi}, \citeyear{9li2020multimodal}). \citet{4palaskar2019multimodal} first release the How2 dataset for open-domain multimodal summarization. \citet{8liu2020multistage} focus on reducing noise in longer text and video summarization and propose a forget gate method to control redundant information. \citet{2li2017multi} and \citet{7khullar2020mast} aim to generate a text summary from more than three modalities: text-video-audio or text-image-audio-video. 
%Most works focus on the way to effectively fuse image feature or the frame in the video into text embedding, they always use cross attention to complete it. 
%So how to represent image features effectively and fuse the different modalities feature is also a challenge.

\noindent\textbf{Multimodal Summarization with Multimodal Output.} Some other works generate a summarization containing both textual and visual content. \citet{11zhu2018msmo} first propose to output a text summary and select the most relevant image. Then \citet{10zhu2020multimodal} add a multimodal loss function to improve the relevance between text summaries and images. Besides, \citet{6li2020vmsmo} and \citet{5fu2020multi} fuse the text-video feature and generate a text summary with a significant image from its associated video. Multimodal output summarization can facilitate readers to obtain crucial information efficiently. Most of these approaches fuse the different modalities of data directly, which neglects the latent correlation and heterogeneity of internal structure among them. Therefore, their performance of summarization is affected and even worse than text summarization.

\section{Problem Formulation}
Our HCSCL model takes a long article and the associated images as the input and generates a summarization containing both the textual summary and the most representative image. $X^t = \left\{ S_1,S_2,...,S_n  \right\}$ is used to denote the textual input, which consists of $n$ sentences, and $X^v = \left\{ I_1,I_2,...,I_p  \right\}$ is used to denote the visual input associated $p$ images. $S_i = \left\{ x_{i,1},x_{i,2},...,x_{i,m} \right\}$ means the sentence $S_i$ has $m$ words. The summarization can be divided into text output and visual output. The textual output is denoted as a sequence of words $Y^t = \left\{ y_1,y_2,...,y_t  \right\}$, and the visual output is $Y^v = \left\{I  \right\}$. In order to generate a multimodal summarization, the model can be formulated as an optimization problem as follows:

$$
\arg\max_{\theta} HCSCL(Y^t,Y^v|X^t,X^v;\theta)
$$

\noindent where $\theta$ is the set of trainable parameters in the model.

\section{Model}
The current multimodal summarization methods have two drawbacks: 1) They neglect the heterogeneity of internal structure between visual content and text. 2) They mainly consider the visual content as the whole, which ignores the hierarchical correlation between different modalities of data.

Therefore, in this work, we propose a hierarchical cross-modality semantic correlation learning model, as shown in Figure \ref{fig:f2}, to improve multimodal summarization. HCSCL comprises three modules: the Modality Feature Encoder is used to encode each modality, the Hierarchical Semantic Correlation Fusion module is used to learn the hierarchical intra- and inter-modality correlation, and the Multimodal Output Summarizer is used to generate the multimodal summary by exploiting the hierarchical correlation. Table 1 presents the key variables used in different modules.

\begin{table}[!h]
\resizebox{84mm}{15mm}{
\centering

\begin{tabular}{p{23mm}ll}
\toprule
                                          & Variable & Description \\\hline
\multirow{2}{*}{\tabincell{l}{Modality Feature \\Encoder}} & $h^0_{i,j}$& The $j^{th}$ word embedding in the $i^{th}$ sentence\\
                                          & $v^0_{i}$  & The $i^{th}$ object feature for the article\\\hline
\multirow{2}{*}{\tabincell{l}{Word-object \\Fusion}}       & $h^1_{i,j}$& The $j^{th}$ word embedding fused with object feature\\
                                          & $v^1_{i}$  & The $i^{th}$ object feature fused with word embedding\\\hline
\multirow{4}{*}{\tabincell{l}{Sentence-scene \\Fusion}}    & $h^2_{i}$  & The $i^{th}$ sentence embedding for the article\\
                                          & $v^2_{i}$  & The $i^{th}$ scene feature for the article\\
                                          & $h^3_{i}$  & The $i^{th}$ sentence embedding fused with scene feature\\
                                          & $v^3_{i}$  & The $i^{th}$ scene feature fused with sentence embedding\\
                                          
\bottomrule
\end{tabular}
}
\caption{Description of key variables.}
\label{table:variable}
\end{table}

\subsection{Modality Feature Encoder}

\quad \textbf{Text Encoder.} We employ LSTM to encode an input article to a vector representation. Specifically, given an input article $X^t=\{S_1, S_2, \cdots, S_n\}$, the embedding of the words in the $i^{th}$ sentence $S_i = \{x_{i,1}, \cdots, x_{i,m}\}$ is learned as follows:\\  
\begin{equation}
   \stackrel{\longrightarrow} {h^0_{i,j}} = LSTM(E[x_{i,j}], \stackrel{\longrightarrow}{h^0_{i-1,j}})
\end{equation}
\begin{equation}
    \stackrel{\longleftarrow}{h^0_{i,j}} = LSTM(E[x_{i,j}], \stackrel{\longleftarrow}{h^0_{i+1,j}})
\end{equation}
where $E[x_{i,j}]$ is the embedding vector of word $x_{i,j}$. $h^0_{i,j}=[\stackrel{\longrightarrow}{h^0_{i,j}}; \stackrel{\longleftarrow}{h^0_{i,j}}]$ represents the hidden state of the $j^{th}$ word in the $i^{th}$ sentence. The $i^{th}$ sentence's word embedding is $\left\{{h^0_{i,1},h^0_{i,1},...,h^0_{i,m}}\right\}$.

% 下标
\ \textbf{Image Encoder.} Given a set of images $X^v=\{I_1, I_2, \cdots, I_p\}$ as the visual input, we apply a Faster R-CNN (\citealp{ren2016faster}) initialized with ResNet-101 (\citealp{he2016deep}) to obtain object proposals for each image. For each object proposal, a triple set $(o_i, l_i, a_i)$ is extracted. $o_i$ is a feature vector extracted from the region of interest (ROI) pooling layer in the Region Proposal Network. $l_i$ is a 4-dimensional bounding box location representation. $a_i$ is a one-hot attribute class feature. Then we use the triple set to form a representation of the $i^{th}$ object as follows:
\begin{equation}
    v_i^0 = Concat(o_i, W^{l}l_i, W^{a}a_i),
\end{equation}
where $W^{l}$ and $W^{a}$ are trainable embedding matrices. $\left\{v_1^0,v_2^0,...,v_q^0\right\}$ is a set of $q$ objects' features.

\subsection{Hierarchical Semantic Correlation Fusion}

Most existing approaches embed the whole image as a global vector, ignoring the internal and cross-modal correlation in the multimodal data. It is not effective to integrate the important information from different modalities. We propose a hierarchical semantic correlation fusion module, which can capture the hierarchical intra- and inter-modal correlation. It learns the important information in each modality by exploiting the intra-modal correlation at different grains and hence learns the important information in the multimodal content by exploiting the inter-modal correlation to reinforce each other, as Figure \ref{fig:f2} shows. This module learns the correlation in two levels: word-object fusion and sentence-scene fusion. The word-object fusion component learns the correlation between words and objects to add visual information into the text entity features. In the sentence-scene fusion component, a visual relation graph is built on the objects to form a sub-scene graph fused with the sentences to learn the inter-modal correlation. Notably, an attention-based cross-modality encoder (CME) (\citealp{tan2019lxmert}) is employed to learn the inter-modality correlation, which enhances the critical information by each other.

% \subsubsection{Token-object Fusion}
\ \textbf{Word-object Fusion.} The framework of CME is shown in Figure \ref{fig:f2}(b). Instead of directly using the attention mechanism, we employ cross-modality attention to generate a fused representation for the multimodal content. CME consists of three parts: cross-attention layer, self-attention layer, and feed-forward layer.

To fuse the $i^{th}$ sentence's word embeddings $\left\{{h^0_{i,1},h^0_{i,1},...,h^0_{i,m}}\right\}$ and objects' features $\left\{v_1^0,v_2^0,...,v_q^0\right\}$, the cross-attention layer is defined as:
\begin{equation}
a_k = score(h^0_{i,j},v_k^0)
\end{equation}
\begin{equation}
\alpha_k = exp(a_k)/ \sum_qexp(a_k)
\end{equation}
\begin{equation}
CrossAtt_{h\rightarrow v} = \sum_q \alpha_kv_k^0
\end{equation}
where $h^0_{i,j}$ is a query word vector, $v_k^0$ is visual object vectors and $score$ is defined as product function to calculate the similarity value $a_i$. The three steps of CME is defined as follows:
\begin{equation}
h^{Cross}_{i,j} = CrossAtt_{h\rightarrow v}(h^0_{i,j},\left\{v_1^0,v_2^0,...,v_q^0 \right\})
\end{equation}
\begin{equation}
h^{Self}_{i,j} = SelfAtt_{h\rightarrow v}(h^{Cross}_{i,j},\left\{h^{Cross}_{i,j} \right\})\\
\end{equation}
\begin{equation}
\left\{h_{i,j}^{out}\right\}= FF({\left\{h^{Self}_{i,1},h^{Self}_{i,2},...,h^{Self}_{i,m} \right\}})
\end{equation}
where $h^{Cross}_{i,j}$ is the result after cross-attention layer, $h^{Self}_{i,j}$ is the result after self-attention and $FF(*)$ is the feed-forward layers. Residual connection and layer normalization are also added after each sub-layer. Three steps are repeated $N_x$ times.

The inter-modal correlation between the textual words and the visual objects is learned by cross-modality attention. The learned multimodal representation is as follows:
\begin{equation}
h^1_{i,j} = CME(h^0_{i,j},\left\{v_1^0,v_2^0,...,v_q^0\right\})
\end{equation}
\begin{equation}
v^1_{i} = CME(v_{q}^0,\left\{h^0_{i,1},h^0_{i,2}...,h^0_{i,m}\right\})
\end{equation}
where $h^1_{i,j}$ is the representation of the $j^{th}$ word in the $i^{th}$ sentence after the fusing procedure, and $v^1_{i}$ is the fused representation of the $i^{th}$ object. $\left\{h^1_{i,1},h^1_{i,2},...,h^1_{i,m}\right\}$ is the word embeddings fused with object features and  $\left\{v_1^1,v_2^1,...,v_q^1\right\}$ is the object features fused with word vectors.

% \subsubsection{Sentence-scene Fusion}
\ \textbf{Sentence-scene Fusion.} After obtaining the fused word features, for each sentence, we use LSTM to obtain a representation of the entire sentence. Then, the sentence representations are further fused with the visual feature by exploiting the correlation. Specifically, the representation of the $i^{th}$ sentence is computed as follows:
\begin{equation}
    h^2_i=LSTM(h^1_{i,1},h^1_{i,2},...,h^1_{i,m}).
\end{equation}

On the other side, a part of objects in an image is correlated, forming a scene to denote a more abstract concept or activity. The scenes are critical components of a summary. We propose an Image Graph Encoder to learn the scene representation, which can capture both the structure and the content information, as shown in Figure \ref{fig:f2}(c). First, based on the bounding box extracted by the image encoder, an Intersection over Union (IOU) score is calculated for every two objects. Next, a relation graph with adjacency matrix $A$ is constructed, where $A_{ij}=1$ if the IOU score exceeds the threshold value and $A_{ij}=0$ else.

Then, we calculate the edge weight as follows. Given two target node feature $v^1_i$ and $v^1_j$, the feature score $s^{feature}_{ij}$ from node $i$ to $j$ is first calculated:

\begin{equation}
s^{feature}_{ij} = w_1^T\sigma(w_2\cdot Concat(v^1_i,v^1_j))
\end{equation}
where $w_1$ and $w_2$ are learned parameters, $\sigma$ is the activation function. For every directed edge of node $i$, we apply a softmax function over the IOU score $s_{ij}^{IOU}$ and feature score $s^{feature}_{ij}$ to obtain the edge weight $s^{edge}_{ij}$:

\begin{equation}
s^{edge}_{ij} = exp(s_{ij}^{IOU}\cdot s^{feature}_{ij})/ \sum_{t\in N(i)}exp(s_{it}^{IOU}\cdot s^{feature}_{it})
\end{equation}
where $N(i)$ is the neighbors of $i$. Next, node feature is updated and combined with the connected node feature to formulate a sub-scene graph representation:

\begin{equation}
\tilde{v}^1_i = \sigma(v^1_i+\sum_{j\in N(i)}s^{edge}_{ij} A_{ij} w_3 v^1_j)
\end{equation}
\begin{equation}
\left\{ v_p^2 \right\} = readout(\tilde{v}^1_1,\tilde{v}^1_2,...,\tilde{v}^1_q)
\end{equation}
where $w_3$ is the learned parameters, $readout$ is a node value aggregation function used to combine nodes' vectors to generate sub-scene graph vectors. $\left\{v^2_1,v^2_2,...,v^2_p\right\}$ is a set of $p$ scene vectors. We define the readout function as:

\begin{equation}
    \left\{ v^2_p \right\} = \frac{1}{N}\sum_{i\in N(i)}\tilde{v}^1_i+Maxpooling({\tilde{v}_1^1,\tilde{v}_2^1,...,\tilde{v}_q^1})
\end{equation}

Finally, the sentence-scene fused features are generated by CME as follows:
\begin{equation}
    h^3_{i} = CME(h_i^2,\left\{v^2_1,v^2_2,...,v^2_p\right\}),
\end{equation}
\begin{equation}
    v^3_{i} = CME(v^2_{i},\left\{h^2_{1},h^2_{2},...,h^2_{n}\right\}).
\end{equation}
where $h^3_{i}$ is the representation of the $i^{th}$ sentence after the fusing procedure, and $v^3_{i}$ is the representation of the $i^{th}$ scene. $\left\{h^3_{1},h^3_{2},...,h^3_{n}\right\}$ is the set of sentence embedding fused with scene graph feature and  $\left\{v_1^3,v_2^3,...,v_p^3\right\}$ is the set of scene graph feature fused with sentence feature.

\subsection{Multimodal Output Summarizer}
The summarizer generates a text summary associated with the most relevant image. In the text summary generation, hierarchical attention is built to combine both word and sentence features. In the image selector,  an object-scene gate mechanism is proposed to select an image as the visual output.

% \subsubsection{Text Summary Generation}
\ \textbf{Text Summary Generation.} In the hierarchical fusion module discussed above, visual objects and scenes features are respectively fused into words and sentences. In this summarizer, a hierarchical attention method is proposed to exploit the two levels of correlation in the decoder stage.

First, LSTM is used to decode sentence features, and the hidden state $h^{\prime}_i$ is generated as follows:

\begin{equation}
    \left\{h^{\prime}_{1},h^{\prime}_{2},...,h^{\prime}_{t}\right\}=LSTM(h^3_{1},h^3_{2},...,h^3_{n})
\end{equation}

For the sentence level, the $i^{th}$ sentence weight $\beta^{sent}_i$ is calculated as follows:

\begin{equation}
    \beta^{sent}_i = \mathop{softmax}\limits_{i}(score(h^3_{i},h^{\prime}_{t-1}))
\end{equation}

For the word level, the weight $\beta^{word}_{i,j}$ of the $j^{th}$ word in the $i^{th}$ sentence is calculated as follows:

\begin{equation}
    \beta^{word}_{i,j} = \mathop{softmax}\limits_{i,j}(\beta^{sent}_{i} \cdot  score(h^1_{i,j},h^{\prime}_{t-1}))
\end{equation}
where $h^3_{i}$ and $h^1_{i,j}$ are sentence and word features respectively. Then the context vector $c_t$ at timestep $t$ and the word probability of $y_t$ are calculated as follows:

\begin{equation}
    c_t = \sum_{i=1}^N\sum_{j=1}^M\beta^{word}_{i,j}h^1_{i,j}
\end{equation}
\begin{equation}
    p(y_t|y_{1:t-1}) = softmax(V^T  FF(h^{\prime}_t,c_t))
\end{equation}
where $V$ is the vocabulary weight matrix. The total loss of textual summary is calculated as follows:
\begin{equation}
    L_{text} = -\sum_{t}\log{p(y_t)}
\end{equation}
%We also follow \citet{see2017get} to improve decoder quality.

% \subsubsection{Image Summary Selector}
\ \textbf{Image Selector.} The most relevant image should match the summary semantic in both the object level and the scene level. Therefore, the summary image is selected based on both the object features and the scenes graph features. First, the hidden state features $h^{\prime}_t$ are used to calculate the relevance score with object features $v^1_i$ and scene features $v^3_j$. Then, an object-scene gate mechanism is proposed to calculate the weights of the two types of features. The score of image $I$ is calculated as follows:
\begin{equation}
    s^{obj}_i = \sigma(FF(v^1_i h^{\prime}_t))
\end{equation}
\begin{equation}
    s^{scene}_j = \sigma(FF(v^3_j h^{\prime}_t))
\end{equation}
\begin{equation}
    \lambda^{*} = \sigma(FF(h^{\prime}_t))
\end{equation}
\begin{equation}
    s^{image}_I = \lambda^{*} \sum_{i \in I} s^{obj}_i+(1-\lambda^{*})\sum_{j \in I}s^{scene}_j
\end{equation}
where $\sum_{i \in I} s^{obj}_i$ is the sum of all scores of the object features in image $I$, $\sum_{j \in I}s^{scene}_j$ is the sum of all scores of the scene graph features in image $I$, and $\lambda^{*}$ is a balance weight between them. The image with the highest score is considered as the image output. The image summary loss is calculated as follows:

\begin{equation}
    L_{image} = -\log({softmax(s^{image}_{I})})
\end{equation}

Finally, the total loss is calculated as follows:
\begin{equation}
    L = L_{text} + {\lambda}L_{image}
\end{equation}

\section{Experimental Setup}

\subsection{Dataset}
% 后面再添数据可从这里改
There is currently one dataset \cite{11zhu2018msmo} for MSMO tasks. However, this dataset is marked only on the test set due to the unsupervised image selection method. It is not suitable for our HCSCL, which needs the image objects information and annotating the most relevant image. Therefore, we construct a large-scale Chinese summarization dataset with complete image annotations and fine-grained object information. Specifically, documents and summaries are selected from TTNews \cite{hua2017overview} and THUCNews \cite{maosongsun2016thuctc}, including sports, entertainment, current politics, society, technology, etc. For each selected document, we search it in Baidu Image Searcher\footnote{https://image.baidu.com/}, and then about ten images with their captions are crawled from the website. Next, we delete noisy images such as gifs, thumbnails, and advertisements.

Image annotation from large amounts of data and images is a time-consuming and laborious task. Since the images we crawled have image captions, we use automatic selection and manual selection methods for image annotation. A Bert \cite{devlin2018bert} and ESIM \cite{chen2016enhanced} semantic matching model pretrained on LCQMC corpus \cite{liu2018lcqmc} is used to infer the relevance of article summary and image caption. Then, for automatic selection, we select the three images with the highest relevance in each document and discard the remaining images. For manual selection, five volunteers are employed to select one of the three images as the most relevant image. In addition, for each image, a Faster R-CNN feature extractor \cite{2017Bottom} is applied to extract the object features, bounding box locations, and attribute classes. Then, the bounding box locations are used to calculate the IOU between objects to generate the adjacency matrices of relation graphs. Finally, a multimodal summarization dataset with complete image annotations and image object features is constructed, including 52656 for training, 5154 for validation, and 5070 for testing. More details about our dataset is illustrated in Table \ref{table:dataset}.
%According to our statistics, the average length of document content and reference summary are 955.26 Chinese characters and 36.61 Chinese characters respectively.

\begin{table}[]
\resizebox{80mm}{15mm}{
\centering
\begin{tabular}{cccc}
\toprule 
                                & Train  & Valid  & Test   \\ \hline
Num. Documents                  & 52656  & 5154   & 5070   \\
Avg. Num. Words in Article     & 953.72 & 956.29 & 970.21 \\
Avg. Num. Sentences in Article  & 19.52  & 19.57  & 19.86  \\
Avg. Num. Words in Summary     & 36.66  & 36.33  & 36.36  \\
Avg. Num. Words in Caption     & 25.07  & 25.10  & 25.10  \\
Avg. Num. Objects in Image      & 7.25   & 7.16   & 7.32   \\
Avg. Num. Scene Graphs in Image & 4.74   & 4.70   & 4.80   \\ \bottomrule
\end{tabular}
}
\caption{Corpus statistics of our dataset.}
\label{table:dataset}
\end{table}

% \begin{figure*}[htbp]
%     \begin{center}
%     \includegraphics[scale=0.55]{case_study.pdf}
%     \end{center}
%     \caption{Pie chart result for human evaluation}
%     \label{fig:case_study}
% \end{figure*}

\begin{table*}[!hpt]
\centering
\resizebox{155mm}{32mm}
{
\begin{tabular}{llllllllll}

\toprule 
\multicolumn{1}{c}{} &
  \multicolumn{1}{c}{R-1} &
  \multicolumn{1}{c}{R-2} &
  \multicolumn{1}{c}{R-L} &
  \multicolumn{1}{c}{B-1} &
  \multicolumn{1}{c}{B-2} &
  \multicolumn{1}{c}{B-3} &
  \multicolumn{1}{c}{B-4} &
  \multicolumn{1}{c}{IP}\\

\hline
\textit{Traditional Textual Model}      & & & & & & & \\
PG \cite{see2017get}   & 42.54 & 28.25 & 39.95 & 38.26 & 30.78 & 24.55 & 19.92 & -  \\
S2S \cite{luong2015effective} & 30.13 & 14.40 & 28.61 & 26.36 & 17.55 & 11.07 & 7.44 & - \\
TextRank \cite{mihalcea2004textrank}        & 22.22 & 10.39 & 18.49 & 28.32 & 8.48 & 4.47 & 2.78 & - \\

\hline
\textit{Multimodal Summarization Model}       & & & & & & & \\
MAtt \cite{12li2018multi}          & 41.30 & 24.70 & 37.55 & 37.41 & 28.44 & 21.13 & 16.02 & -  \\
HOW2 \cite{4palaskar2019multimodal}           & 39.53 & 20.28 & 35.55 & 35.44 & 24.42 & 17.81 & 13.71 & -  \\
MSE \cite{9li2020multimodal}     & 42.99 & 28.78 & \textbf{41.79} & 39.02 & 31.51 & 25.21 & 20.50 & -  \\
\hline
\textit{Multimodal Summarization Output Model}      & & & & & & & \\
MSMO \cite{11zhu2018msmo}            & 42.89 & 28.25 & 39.86 & 39.06 & 31.26 & 24.78 & 19.93 & 33.50 \\
MOF \cite{10zhu2020multimodal}           & 42.41 & 28.10 & 39.60 & 38.04 & 30.55 & 24.38 & 19.71 & 33.16 \\
VMSMO \cite{6li2020vmsmo}           & 42.68 & 28.35 & 41.34 & 38.75 & 31.20 & 24.75 & 19.99 & 32.86 \\
\hline
\textit{our models}     & & & & & & & \\
HCSCL text output only            & 42.11 & 27.57 & 39.52 & 37.90 & 30.26 & 23.84 & 19.13 & - \\
HCSCL multimodal output          & \textbf{43.64} & \textbf{29.00} & 40.94 & \textbf{39.64} & \textbf{31.91} & \textbf{25.40} & \textbf{20.54} & \textbf{40.98} \\
\bottomrule
\end{tabular}
}
\caption{Rouge, BLEU and IP scores comparison with summarization baselines.}
\label{table:1}
\end{table*}

\subsection{Baseline models}
% 时间充裕可以多做几个实验
% 具体描述可以改一改 有的是直接抄的
We compare our model in three categories of baselines, a total of nine models.\par

\textit{{Multimodal Summarization Output Models. }}\noindent\textbf{MSMO} \cite{11zhu2018msmo} is the first model on multimodal summarization with multimodal output task, which applies attention to text and images during generating textual summary and uses coverage mechanism to select image. \textbf{MOF} \cite{10zhu2020multimodal} is the model based on MSMO, which considers image accuracy as another loss. \textbf{VMSMO} \cite{6li2020vmsmo} is a video-based news summarization model, which proposes a Dual-Interaction-based Mutilmodal Summarizer. \par

\textit{Multimodal Summarization Models. }\noindent\textbf{MAtt} \cite{12li2018multi} is an attention-based model, which utilizes image filtering to select visual content to enhance sentence features. \textbf{HOW2} \cite{4palaskar2019multimodal} the first model proposed to generate a textual summary by exploiting video content. \textbf{MSE} \cite{9li2020multimodal} is a model which focuses on enhancing the ability of the encoder to identify highlights of the news.\par
%and utilizes a selective gate network

\textit{Traditional Textual Models. }\noindent\textbf{PG} \cite{see2017get} is a sequence-to-sequence framework combined with attention mechanism and pointer network. \textbf{S2S} \cite{luong2015effective} is a standard sequence-to-sequence architecture using a RNN encoder-decoder with a global attention mechanism. \textbf{TextRank} \cite{mihalcea2004textrank} is a graph-based extraction summarizer which represents sentences as nodes and uses edges to weight similarity.\par

\subsection{Evaluation Metrics}
% rouge还要有图片评价
As a widely used evaluation metric in text summarization, ROUGE (\citealp{lin2004rouge}) and BLEU (\citealp{papineni2002bleu}) are applied to evaluate the quality of the generated text summary. Besides, image precision (IP) is used to evaluate the quality of the output image (\citealp{11zhu2018msmo}), which is as follows:
%ROUGE-1, ROUGE-2 and ROUGE-L respectively represent the F1-scores of unigram, bigram and the longest common subsequence.
\begin{equation}
    IP = \frac{1}{N}\sum\limits_{i}I(ann^{img}_i=rec^{img}_i)
\end{equation}
where $ann^{img}$ and $rec^{img}$ denote the annotated image and the output image respectively.

\subsection{Implementation Detail}
% 做完实验再写
The experiments are conducted in Pytorch on NVIDIA Tesla V100 GPUs. We freeze the basic version of pre-trained Bert-base-Chinese \cite{2018BERT} for the original text embedding, which has 12 layers, 12 heads, 768 hidden dimensions, and 110M parameters. The Faster R-CNN feature extractor \cite{2017Bottom} is used for image object detection, and objects with confidence greater than 0.55 are selected. Each object has a 2048-dimensional representation. The IOU threshold is set to 0.2. To train the model, we employ Adam \cite{2014Adam} as the optimizer, and the batch size is set at 16. The initial learning rate is set to $5e^{-4}$ and multiply by 0.8 every six epochs.
\section{Results and Analysis}

\subsection{Overall Performance}
To validate the effectiveness of our model, we compare our model with three kinds of baselines: Traditional Textual Model, Multimodal Summarization model, and Multimodal Summarization Output Model. Table \ref{table:1} shows the result of comparative models on the dataset. We can obtain several conclusions from this table. 
First, it shows that HCSCL achieves state-of-the-art performance in almost all evaluation metrics. HCSCL outperforms baselines 1.51\%, 0.76\% in terms of Rouge-1, Rouge-2 and 1.48\%, 1.27\%, 0.75\%, 0.20\% in terms of BLEU-1, BLEU-2, BLEU-3, BLEU-4 and 22.32\% in IP. It demonstrates the superiority of the hierarchical cross-modality correlation learning model. By exploiting the intra-modality correlation learning and the inter-modality feature aligning, the visual content can reinforce a specific part of the representation of text content, and the text content can reinforce to select the relevant image. 
Second, in Rouge-L, HCSCL (40.94) is slightly worse than the text output model MSE (41.79). After analyzing the cases, we find that the text outputs of HCSCL are more relevant to image semantics. Therefore, the longest common subsequence (R-L) is less matched with the text description, affecting this metric's evaluation.

\subsection{Ablations Study}
Four comparative experiments are designed to verify the effectiveness of word-object fusion, sentence-scene fusion, and hierarchical structure. The result is shown in Table \ref{table:2}. We can see that with the hierarchical structure, the model has better performance. With the word-object fusion, the R-1, R-2, and B-1 are higher. With the sentence-scene fusion, the image selection metric IP is higher. However, when skipping the low-level word-object fusion module and using the sentence-scene fusion module directly, the model's performance worsens text evaluation metrics. It demonstrates that the low-level semantic correlation helps learn the high-level semantics and improve the summarization quality.

\begin{table}
\resizebox{84mm}{!}{
\begin{tabular}{lllllll}
\toprule 
\multicolumn{1}{c}{} &
  \multicolumn{1}{c}{R-1} &
  \multicolumn{1}{c}{R-2} &
  \multicolumn{1}{c}{B-1} &
  \multicolumn{1}{c}{IP}\\
\hline
HCSCL Word-Object only   & 41.70  & 25.37 & 38.01  & 32.31    \\
HCSCL Sentence-Scene only & 38.99  & 20.41  & 35.58  & 33.36  \\
\hline
HCSCL \emph{w/o} Sentence-Scene Fusion          & 41.05 & 24.04  & 37.44  & 31.94   \\
HCSCL \emph{w/o} Word-Object Fusion           & 38.35  & 19.94  & 34.97  & 30.83   \\
\hline
HCSCL        & 43.64  & 29.00   & 39.64  & 40.98  \\
\bottomrule
\end{tabular}
}
\caption{The ablation results for HCSCL. “Word-Object only” means a model only contains Text Encoder, Image Encoder, and Word-Object Fusion. “Sentence-Scene only” means a model only contains Sentence Encoder, Image Graph Encoder, and Sentence-Scene Fusion.}
\label{table:2}
\end{table}

\subsection{Fine-grained Semantic Diversity Analysis}

\begin{figure}[t]
    \begin{center}
    \includegraphics[scale=0.50]{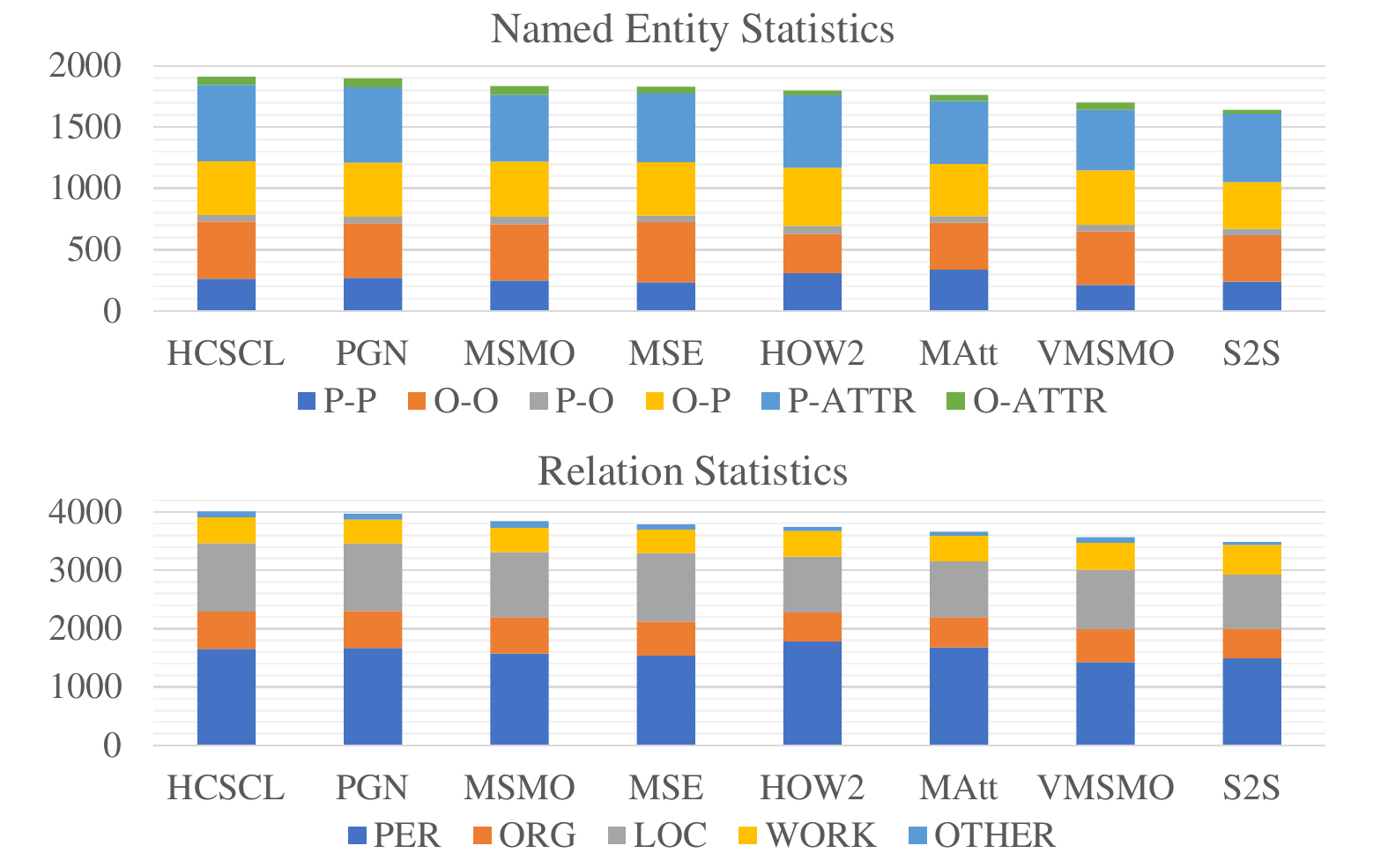}
    \end{center}
    \caption{The histogram statistics of the number of named entities and the relations in output text summary.}
    \label{fig:relation}
\end{figure}

To test fine-grained degree \cite{yang2020hierarchical}, we calculate the number of named entities and relations in the output text summaries. The pre-trained relation extraction model NEZHA \cite{wei2019nezha} is trained on the LIC2021 Competition dataset\footnote{https://aistudio.baidu.com/aistudio/competition/detail/65}, which extracts 24 types of name entities and 42 types of relations in our dataset. Then, the text summaries are input to the relation extraction model. Next, 24 name entities are divided into five categories: PER, ORG, LOC, WORK and OTHER, and 42 relations are divided into six categories: P-P, O-O, P-O, O-P, P-ATTR, and O-ATTR.
% \footnote{The meaning of abbreviations can be found in Appendix} 
The statistic result of the experiment is shown in Figure \ref{fig:relation}. Note that when the ROUGE is similar or slightly lower (as shown in Table \ref{table:1}), PGN can extract more name entities and relations than the other multimodal summarization models, which shows that the summaries generated by the previous multimodal summarization models have a certain loss of semantic information. Compared with the baselines, the summaries generated by HCSCL have the largest number of named entities and relations. From these results, we find that HCSCL can generate summaries with the richest semantic information. This is because HCSCL learns the correlation between the objects in the multimodal data, which is more effective in discovering the relationships and entities even if some critical information is missed in the text data.

\begin{table}[]
\resizebox{84mm}{6mm}{
\begin{tabular}{ccccccccc}
\toprule
              & HCSCL & PGN & MSMO & MSE & HOW2 & MAtt & VMSMO & S2S \\ \hline
Ent. & 4006 & 3972 & 3844 & 3778 & 3746 & 3691 & 3566 & 3489 \\ \hline
Rel.      & 1911 & 1900 & 1836 & 1814 & 1801 & 1764 & 1702 & 1643 \\ \bottomrule
\end{tabular}
}
\caption{The total number of name entities and relations in output text summaries.}
\end{table}

\begin{figure}[!htpb]
    \begin{center}
    \includegraphics[scale=0.34]{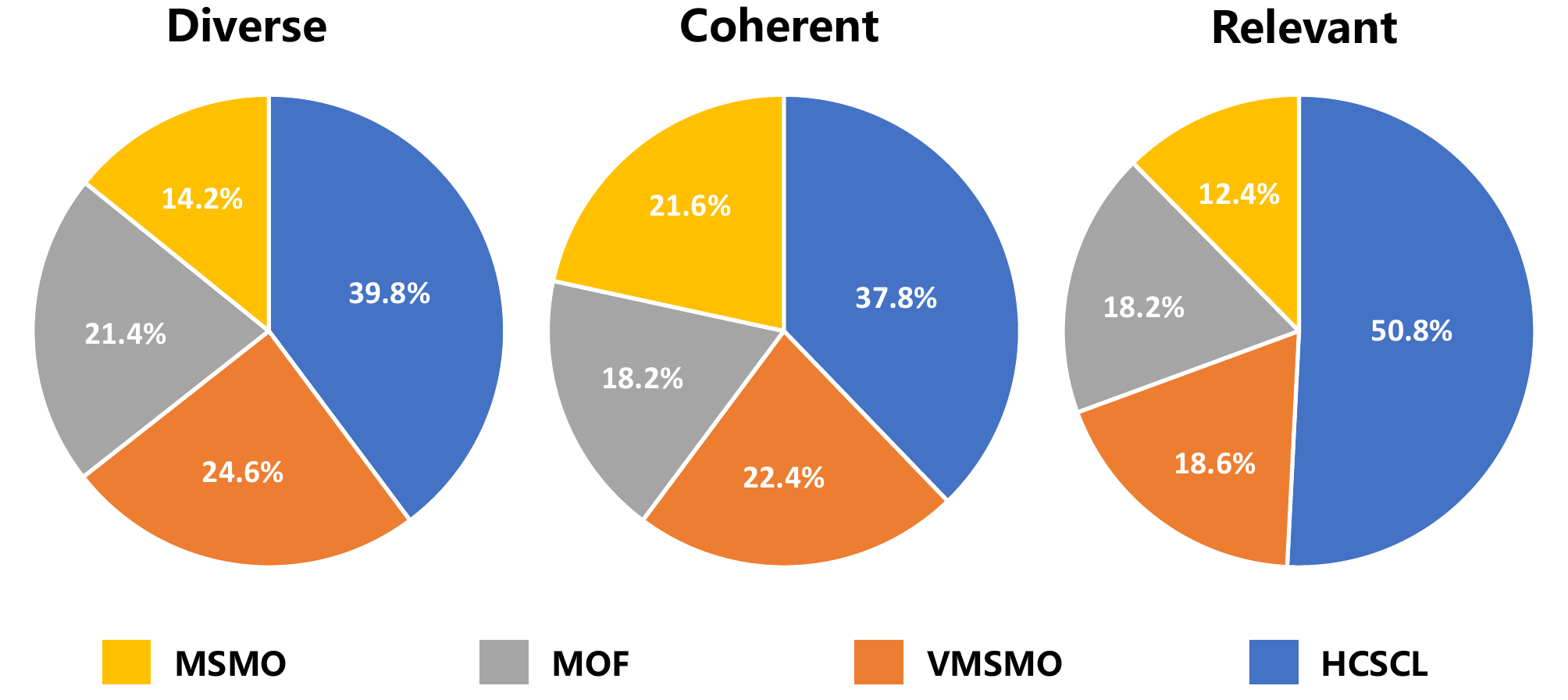}
    \end{center}
    \caption{The pie charts for human evaluation.}
    \label{fig:human evaluation}
\end{figure}

% \begin{figure}
%     \begin{center}
%     \includegraphics[scale=0.28]{img/figure3_v1.pdf}
%     \end{center}
%     \caption{The radar charts of the number of named entities and the relations in output text summary. PER entities include entertaining figures, historical figures, and figures. ORG entities include enterprises/brands, schools, enterprises, and institutions. LOC entities include scenic spots, countries, locations, administrative regions， and cities. WORK entities include songs, TV variety shows, film and television works, book works, literary works, and music albums. OTHER entities include date, number, and award. P-P relations include wife, husband, mother, and father. O-O relations include city, capital, headquarters location, production company, and album. P-O relations include awards, acting, and dubbing. O-P relations include spokesperson, chairman, singer, guest, host, principal, director, protagonist, founder, author, lead actor, and sreenwriter. P-ATTR relations include ancestral home, Dynasty, Graduate School, and nationality. O-ATTR relations include floor area, box office, establishment date, altitude, release time, registered capital, theme song, area, and population.}
%     \label{fig:xingtu}
% \end{figure}
% P-P: The relationship between people. O-O: The relationship between objects. P-O:The relationship between people and objects.P-ATTR:People's attribute.O-ATTAR:Objects' attribute. 
% "P", "O", "ATTR" respectively represent people, object and attribute. 

\subsection{Human Evaluation}
To further evaluate our model's performance, 100 multimodal summaries generated by MSMO, MOF, VMSMO, and HCSCL are selected for human evaluation. Then five graduate students are volunteered to choose the most Diverse, Coherent, and Relevant (how the image matches the textual summary). The evaluation is shown in Figure \ref{fig:human evaluation}. Compared with MSMO, MOF, and VMSMO, the multimodal outputs generated by HCSCL are the most diverse, coherent, and relevant, obtaining 39.8\%, 37.8\%, and 50.8\% votes. It further demonstrates that HCSRF can learn the most representative and abundant semantic information from the article.

% \begin{figure}[t]
%     \begin{center}
%     \includegraphics[scale=0.85]{img/case_study_v1.pdf}
%     \end{center}
%     \caption{Examples of the generated summary by baselines and HCSRF.}
%     \label{fig:case_study}
% \end{figure}

% \subsection{Case Study}
% A case study is shown in Figure \ref{fig:case_study}, which shows a text summary and the relevant image generated by three models. We also show the original article and reference summary. The result shows that HCSCL can generate a fluent and high-quality textual summary and select the most relevant image, which further demonstrate the superiority of HCSCL.

\section{Conclusion}
In this paper, we propose a Hierarchical Cross-Modality Semantic Correlation Learning Model for multimodal summarization, which exploits the intra- and inter-modality correlation in the multimodal data to learn the summary information from both modalities complementarily. The experimental results on the well-designed dataset show that our model can generate the most diverse and coherent text summary with the most relevant image. The novelty of this work is to tackle the multimodal summarization by proposing a multimodal model to learn the heterogeneous structure of different modalities, and hence the correlation between them. This complements the current research, which is not effective to mine the latent and important information inside the multimodal content. 
% \section{Acknowledgements}

% Use \bibliography{yourbibfile} instead or the References section will not appear in your paper
\bibliography{aaai22}

% \bigskip

\end{document}